\documentclass{ieeeaccess}
\pdfoutput=1
\usepackage{cite}
\usepackage{amsmath,amssymb,amsfonts}
\usepackage{graphicx}
\usepackage{textcomp}
\usepackage{changes}

\usepackage{amsmath,bm}
\usepackage{algorithm}
\usepackage{algorithmicx}
\usepackage{algpseudocode}
\usepackage{array}

\usepackage{multirow}
\usepackage{booktabs}
\usepackage{color}
\usepackage{ulem}

\usepackage{verbatim}
\usepackage{amssymb}
\usepackage{wasysym}

\def\BibTeX
{{\rm B\kern-.05em{\sc i\kern-.025emb}\kern-.08emT\kern-.1667em\lower.7ex\hbox{E}\kern-.125emX}}

\begin{document}
\history{}
\doi{}

\title{MV-C3D: A Spatial Correlated Multi-View 3D Convolutional Neural Networks}
\author{\uppercase{Qi Xuan}\authorrefmark{1,2,3},
        \uppercase{Fuxian Li}\authorrefmark{2},
        \uppercase{Yi Liu\authorrefmark{4}, and Yun Xiang\authorrefmark{2}}.}

\address[1]{Institute of Cyberspace Security, Zhejiang University of Technology, Hangzhou 310023, China  }
\address[2]{College of Information Engineering, Zhejiang University of Technology, Hangzhou 310023, China }
\address[3]{Zhejiang Lab, Hangzhou 311121, China }
\address[4]{Institute of Process Equipment and Control Engineering, Zhejiang University of Technology, Hangzhou 310023, China}

\corresp{Corresponding author: Yun Xiang (e-mail: xiangyun@zjut.edu.cn).}

\tfootnote{This work is partially supported by National Natural Science Foundation of China (61572439, 11505153), Zhejiang Provincial Natural Science Foundation of China under (LR19F030001, LY18F030021), and the Key Technologies, System and Application of Cyberspace Big Search, Major Project of Zhejiang Lab (2019DH0ZX01).}

\markboth
{Qi Xuan \headeretal: Multi-view Based 3D Convolutional Neural Networks for 3D Object Classification}
{Qi Xuan \headeretal: Multi-view Based 3D Convolutional Neural Networks for 3D Object Classification}

\begin{abstract}
	As the development of deep neural networks, 3D object recognition is becoming increasingly popular in computer vision community. Many multi-view based methods are proposed to improve the category recognition accuracy. 
	These approaches mainly rely on multi-view images which are rendered with the whole circumference. 
	In real-world applications, however,  3D objects are mostly observed from partial viewpoints in a less range. 
	Therefore,  we propose a multi-view based 3D convolutional neural network, which takes only part of contiguous multi-view images as input and can still maintain high accuracy.
	Moreover, our model takes these view images as a joint variable to better learn spatially correlated features using 3D convolution and 3D max-pooling layers. 
	Experimental results on ModelNet10 and ModelNet40 datasets show that our MV-C3D technique can achieve outstanding performance with multi-view images which are captured from partial angles with less range. The results on 3D rotated real image dataset MIRO further demonstrate that MV-C3D is more adaptable in real-world  scenarios.
	The classification accuracy can be further improved with the increasing number of view images.
\end{abstract}


\titlepgskip=-15pt
\maketitle

\Figure[t!](topskip=0pt, botskip=0pt, midskip=0pt)[width=1\linewidth]{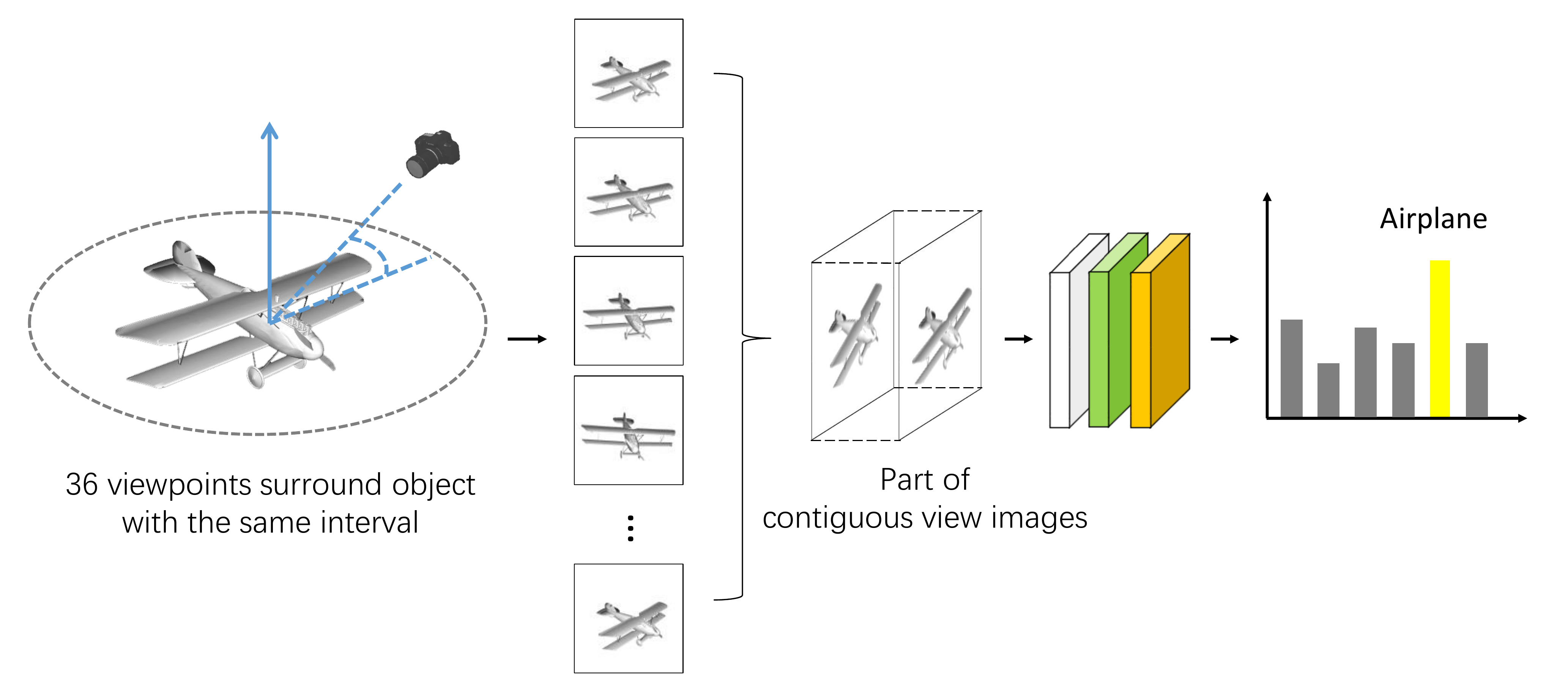}
{Object classification pipeline. We render 3D model into images at 36 predefined viewpoints which are around object as a circumference. The interval between each viewpoint is 10 degree. Our MV-C3D model takes these contiguous images from partial angles in less range as input to predict category label.\label{pipe}}

\section{Introduction}
\label{intro}
\PARstart{R}{ecently}, deep learning technologies have been widely applied to several industrial manufacturing processes~\cite{xuan2018multi, liu2017flame, liu2018ensemble, xuan2018automatic, xuan2018evolving}. The development of convolutional neural networks (CNNs) has enabled the dramatic progress of 3D object recognition technologies. 3D object recognition has a wide range of applications, e.g., automatic driving~\cite{simon2018complex}, robots~\cite{tsarouchi2016method}, and civil monitoring etc~\cite{ji20133d}. In this work, we have proposed a novel 3D CNN architecture which only requires multiple images from limited viewpoint and can still achieve satisfied classification results.

Currently, most CNN architectures are designed specifically for 2D images~\cite{hu2018squeeze}. Therefore, to perform classifications for 3D models, we need to transform the current models based on voxels or 2D images. For voxel based approaches, 3D models are organized as volumetric occupancy grids~\cite{maturana2015voxnet, sedaghat2016orientation, zhi2018toward}. The main advantage of voxel representation is that it can maintain full geometrical information of the original 3D objects. However, those approaches also suffer from the problems of resolution loss and exponentially growing computational cost~\cite{qi2016volumetric}.

Previously, researchers developed multi-view based methods, which can derive comparable results with much lower computational cost~\cite{su2015multi, kanezaki2018rotationnet, wang2017dominant}. However, the multi-view based methods require multiple images derived from various predefined viewpoints in the whole circumference, which is quite impractical for real-world applications. Thus, it is much more desirable to perform successful 3D recognition from multi-view images in limited viewpoints. The existent multi-view image approaches~\cite{su2015multi, wang2017dominant} treat each multi-view image as an independent variable and feed the images into 2D CNNs separately, and the final classifications are derived by aggregating the feature vectors with view-pooling or clustering. Those approaches can easily lead to inferior results by neglecting the spatial correlations between the multi-view images.

Thus, to address these problems, we propose a multi-view based 3D CNN, or MV-C3D.  As shown in FIGURE~\ref{pipe}, our technique takes the multi-view images of objects as input and predicts the corresponding category labels. Unlike the existent multi-view based methods, our model uses multi-view images from only partial angles with less range, which makes it more adaptable in real-world applications. Moreover, our technique considers different viewpoint images as a joint variable instead of independent variables. With the help of 3D convolution and 3D max-pooling layers, our MV-C3D architecture can take advantage of spatial correlations between multi-view images to learn distinguishing features from different objects.

The main contributions of this work are summarized as follows.

\begin{enumerate}
\item We propose the novel multi-view based 3D convolution neural network for the first time, namely MV-C3D, which only requires partial multi-view images from limited viewpoint and outperforms the current multi-view based state-of-the-art classification performance on ModelNet benchmark.

\item We combine the images of different view as a joint variable to learn spatial correlated features by using 3D convolution and 3D max-pooling layers. The visualization of feature maps shows that our network can focus on the same part of object in different view images.

\item We demonstrate experimentally that MV-C3D can get higher classification accuracy with contiguous and increasing view images in partial angles with less range.

\item We test MV-C3D with a 3D rotated real image dataset MIRO with multiple images which was captured from arbitrary but contiguous viewpoint to demonstrate the performance of real-world scenarios.

\end{enumerate}

The experimental results show that, on ModelNet dataset, our proposed architecture can outperform the state-of-the-art multi-view based method MVCNN~\cite{su2015multi} by 3.8\% , multi-modal based method Spherical Projections~\cite{cao20173d} by 0.6\% and DSCNN~\cite{wang2017dominant} by 1.7\%, with the same input modality, respectively.

The rest of the paper is organized as follows. Section \ref{related} provides a detailed review of the related works. Section \ref{method} describes our proposed MV-C3D architecture. Section \ref{experiment} presents the experimental setup and results. Section \ref{conclusion} concludes the paper and discuss future works.

\section{Related Work}
\label{related}
Previously, researchers mainly rely upon local or global descriptors which can map 3D shape information into feature vectors~\cite{guo2013rotational, tang20173d, tombari2010unique, knopp2010hough}. With the breakthrough of CNNs, neural network based approaches are becoming more and more popular. The current existent works can be generalized into two categories: voxel based methods and multi-view images based methods.

\subsection{Voxel based method}

Wu et al.~\cite{wu20153d} constructed a five-layer 3D convolutional deep belief network (CDBN), namely 3D ShapeNet, to learn the probability distribution of 3D voxel grids. Sedaghat et al.~\cite{sedaghat2016orientation} considered 3D object classification as a multi-task problem by introducing object orientation prediction. This model achieved excellent performance, which demonstrates that orientation is also an important aspect for 3D object classification~\cite{maturana2015voxnet, wu2016learning}.

\subsection{Multi-view images based method}
2D images based methods are also important for 3D object classification problem. Su et al.~\cite{su2015multi} proposed a multi-view CNN (MVCNN) based technique to aggregate multiple images  into concise descriptors in a view pooling layer, which lies in the middle of a 2D CNN framework pre-trained on ImageNet~\cite{deng2009imagenet}. Multi-view images are also used in 3D object retrieval applications~\cite{bai2016gift}. Qi et al.~\cite{qi2016volumetric} conducted a comprehensive study on the voxel based and multi-view based CNNs for 3D object classification. According to these works, there are two important factors affecting the model performance: architecture and volume resolution. Therefore, two distinct volumetric networks and multi-resolution filtering technique are proposed. In particular, Feng et al.~\cite{feng2018gvcnn} proposed a group-view convolutional framework which is composed of a hierarchical view-group-shape architecture for correlation modeling towards discriminative 3D shape description. Currently, the state-of-the-art method is~\cite{wang2017dominant}, called DSCNN, which can learn feature vectors from multiple views by using a recurrent cluster strategy. In addition to the above methods, a novel method which can directly work on point cloud data attracts increasing attention~\cite{li2018so, charles2017pointnet}, but the performance is still worse than multi-view images based approaches.

\Figure[b](topskip=0pt, botskip=0pt, midskip=0pt)[width=0.999\linewidth]{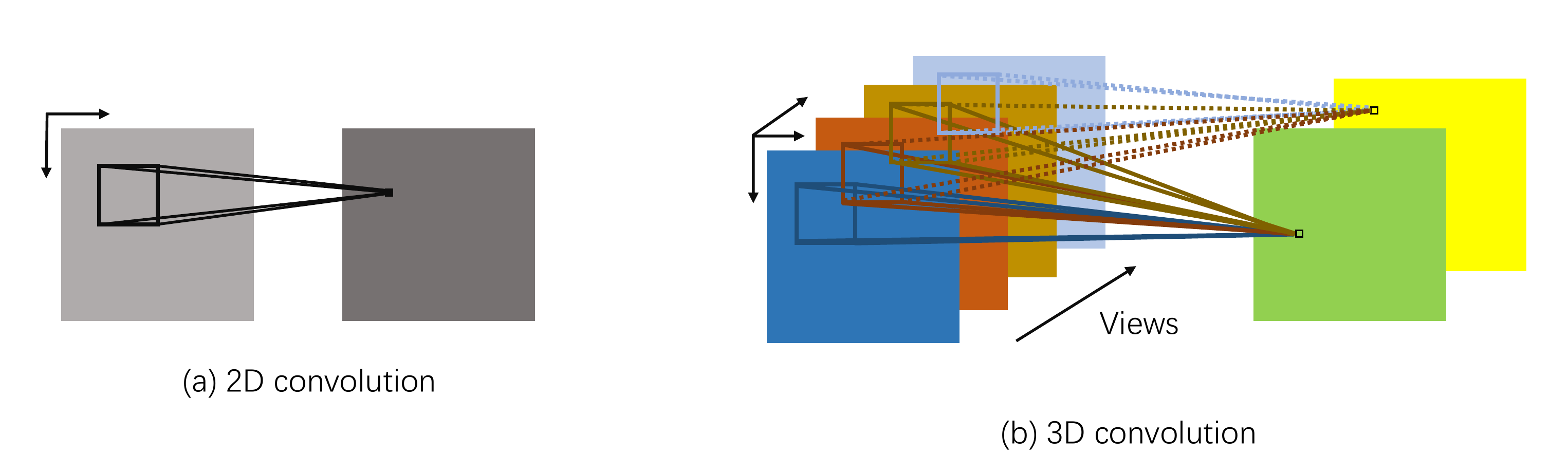}
{Comparison of 2D and 3D convolutions: (a) 2D convolution and (b) 3D convolution. In (b) the kernel size is $3 \times H \times W$, which means that it computes the related features between 3 contiguous views.\label{2d-3d}}

Specifically, MVCNN~\cite{su2015multi} treats each view image as an independent variable and feeds it into 2D CNNs to compute feature maps. Then it directly performs a full stride channel-wise max pooling on these feature maps to generate a unified feature vector. Although MVCNN achieves great success in peral classificaiton, this operation may destroy the spatial correlation information and thus could be further improved. Moreover, DSCNN~\cite{wang2017dominant} uses a clustering strategy which may also cause the loss of viewpoint dimension information. Different from these existent methods, our MV-C3D technique treats multi-view image as a joint variable, and uses 3D convolution and 3D max-pooling to learn both the spatial features and the intrinsic correlations among multi-view images simultaneously.

\section{MV-C3D Method}
\label{method}

\subsection{Multi-view based 3D convolution}
\label{3dconv}

In 2D CNN applications, we only need to compute features from the 2D spatial dimensions and thus, a single-view image of object is sufficient. However, for the 3D object recognition problem, it is required to encode 3D object information from 3D spatial dimensions where different viewpoint images are considered as the third dimension. Compared to 2D CNNs, 3D CNNs can be more efficient and accurate  for multi-view feature learning. In 3D CNNs, 3D convolution is performed by applying a 3D kernel in the view images. FIGURE~\ref{2d-3d} shows the difference between 2D and 3D convolutions. 2D convolution kernel is applied on an image in 2D spatial directions. Thus, it cannot include 3D view information. On the other hand, 3D convolution kernel can preserve spatial correlation information between different view images. Moreover, unlike voxel-based 3D CNNs which focus on learning geometrical features, multi-view based 3D CNNs can capture the correlated features between multi-view images. Formally, the value at position $(x, y, v)$ on the $n$th feature map in the $i$th layer $f_{n}^{i}(x, y, v)$ is given by: 
\begin{equation}
f_{n}^{i}(x, y, v) =  ReLU(b_{n}^{i}+ \sum_{m} h ),
\end{equation}
\begin{equation}
h = \sum_{\bar{x}=0}^{X_{i}-1} \sum_{\bar{y}=0}^{Y_{i}-1} \sum_{\bar{v}=0}^{V_{i}-1}  w_{mn}^{i} (\bar{x},\bar{y},\bar{v}) f_{m}^{i-1} (x+\bar{x},y+\bar{y},v+\bar{v})),
\end{equation}
where $b_{n}^{i}$ is the bias, $h$ is the result of convolution with the $m$th feature map, $(X_{i}, Y_{i}, V_{i})$ is the size of 3D convolution kernel, $(\bar{x}, \bar{y}, \bar{v})$ is the offset, $V_{i}$ is the viewpoint dimension, $w_{mn}^{i} $ is the kernel connected to the $m$th feature map in the $(i-1)$th layer, and $f_{m}^{i-1}$ is the $m$th feature map in the $(i-1)$th layer.

\subsection{Partial multi-view images setup}
\label{input}
Typically, 3D models in online repository are stored as polygon meshes, which are collections of vertices, edges, and faces that define the shape of a polyhedral object. We employ the Phong reflection model~\cite{phong1975illumination} to render 3D models at different predefined viewpoints. We assume that the 3D objects are upright oriented along with $z$-axis~\cite{feng2018gvcnn, wang2017dominant}. As shown in FIGURE~\ref{partial}, we fix the $z$-axis as the rotation axis and then place viewpoints separated by angle $\theta=10^\circ$ around the axis. The viewpoints are elevated by $\phi=30^\circ$ from the ground plane. As a result, we generate 36 view images. Unlike the existing omnibearing viewpoints based methods, only a portion of contiguous images from limited viewpoints are required.

Moreover, 2D images with larger resolution can reserve more information, which can lead to a better performance at the cost of computational time. To balance the computational cost and performance, we set the size of each image to $112\times112$.

\Figure[h](topskip=0pt, botskip=0pt, midskip=0pt)[width=.999\linewidth]{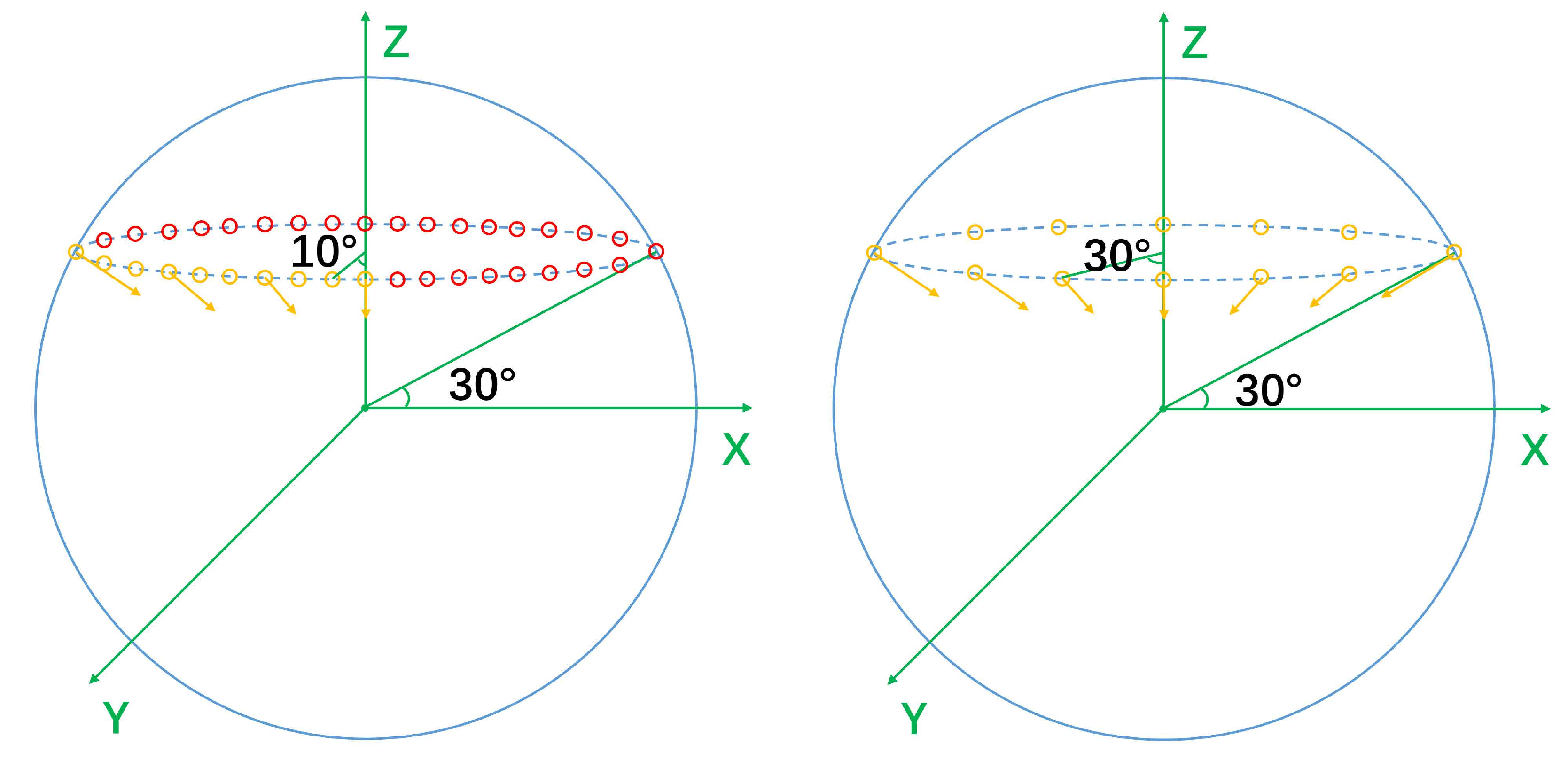}
{Different input representation setup.\label{partial}}

\subsection{Network Architecture}
\label{architecture}

Based on the structure of VGGNet~\cite{simonyan2014very}, we propose MV-C3D, which is essentially a 3D CNN capable of processing contiguous multi-view images (FIGURE~\ref{arch}). The input is a stack cube of multi-view RGB images with the size of $ N\times112\times 112$, where $N$ is the number of views, $112 \times 112$ is the height and width of a single image. Unlike the existent methods~\cite{su2015multi, wang2017dominant}, we do not compute 2D spatial features on different view images independently. Instead, we take these images as an entire instance and learn spatially correlated features between multi-view images.

Our network architecture contains eight 3D convolution layers, five 3D max-pooling layers, three fully-connected layers, and a softmax function to estimate the output distribution. For the fully-connected layers, the dimensions of the first two layers are equal to 4096, while that of the third layer is determined by the number of classes. The activation function of the 3D convolution and fully-connected layers is rectified linear units (ReLUs). We also implement a dropout layer following the first two fully-connected layers to reduce overfitting.

\Figure[t!](topskip=0pt, botskip=0pt, midskip=0pt)[width=1\linewidth]{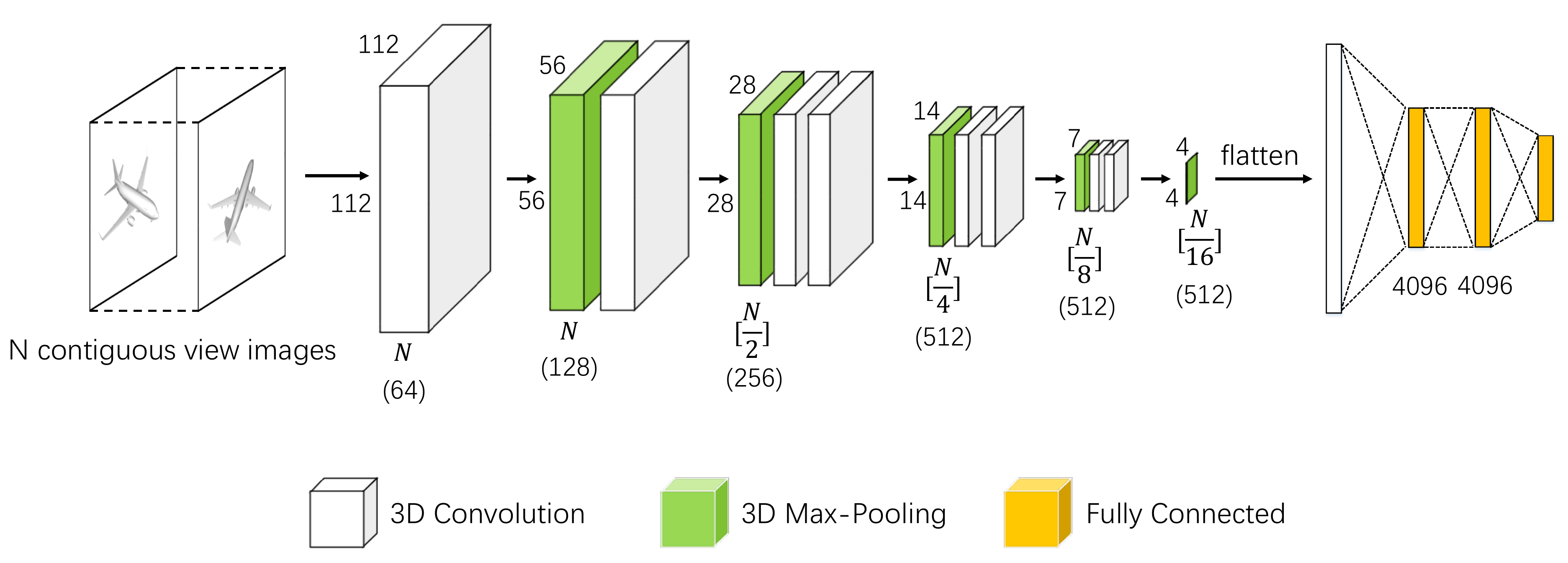}
{The architecture of our MV-C3D. $N$ is the number of view images. The white layer represent the 3D convolution operation, green layer represent 3D max-pooling, and yellow layer represent fully connected layers. $\left [ \cdot  \right ]$ is the floor function, and the (number) represents the channels of feature cube.\label{arch}}

\begin{table*}[!t]
	\centering
	\caption{Size for each layer of MV-C3D.}
	\label{para}
	\begin{tabular}{cccc}
		\toprule
		\multirow{2}{*}{\textbf{Name}} & \multirow{2}{*}{\textbf{Type}} & \multirow{2}{*}{\textbf{Filter size/stride}} & \multirow{2}{*}{\textbf{Output size}} \\
	    & & &\\
	    \midrule
		&&&\\
		Input     &                 &         & $3 \times N \times 112 \times 112$    \\ [5pt]
		Conv1     & Convolution     & $v \times 3 \times 3 / 1 \times 1 \times 1$ & $N \times 112 \times 112 \times 64 $  \\[5pt]
		Pool1     & Max pooling     & $1 \times 2 \times 2 / 1 \times 2 \times 2$ & $N \times 56 \times 56 \times 64 $    \\[5pt]
		Conv2     & Convolution     & $v \times 3 \times 3 / 1 \times 1 \times 1$ & $N  \times 56 \times 56 \times 128 $ \\[5pt]
		Pool2     & Max pooling     & $2 \times 2 \times 2 / 2 \times 2 \times 2$ & $\left [ \frac{N}{2} \right ] \times 28 \times 28 \times 128 $ \\[5pt]
		Conv3\_a  & Convolution     & $v \times 3 \times 3 / 1 \times 1 \times 1$ & $\left [ \frac{N}{2} \right ] \times 28 \times 28 \times 256 $ \\[5pt]
		Conv3\_b  & Convolution     & $v \times 3 \times 3 / 1 \times 1 \times 1$ & $\left [ \frac{N}{2} \right ] \times 28 \times 28 \times 256 $ \\[5pt]
		Pool3     & Max pooling     & $2 \times 2 \times 2 / 2 \times 2 \times 2$ & $\left [ \frac{N}{4} \right ] \times 14 \times 14 \times 256 $ \\[5pt]
		Conv4\_a  & Convolution     & $v \times 3 \times 3 / 1 \times 1 \times 1$ & $\left [ \frac{N}{4} \right ] \times 14 \times 14 \times 512 $ \\[5pt]
		Conv4\_b  & Convolution     & $v \times 3 \times 3 / 1 \times 1 \times 1$ & $\left [ \frac{N}{4} \right ] \times 14 \times 14 \times 512 $ \\[5pt]
		Pool4     & Max pooling     & $2 \times 2 \times 2 / 2 \times 2 \times 2$ & $\left [ \frac{N}{8} \right ] \times 7 \times 7 \times 512 $ \\[5pt]
		Conv5\_a  & Convolution     & $v \times 3 \times 3 / 1 \times 1 \times 1$ & $\left [ \frac{N}{8} \right ] \times 7 \times 7 \times 512 $ \\[5pt]
		Conv5\_b  & Convolution     & $v \times 3 \times 3 / 1 \times 1 \times 1$ & $\left [ \frac{N}{8} \right ] \times 7 \times 7 \times 512 $ \\[5pt]
		Pool5     & Max pooling     & $2 \times 2 \times 2 / 2 \times 2 \times 2$ & $\left [ \frac{N}{16} \right ] \times 4 \times 4 \times 512 $ \\[5pt]
		Fc1 & Fully connected & & $4096$   \\[5pt]
		Fc2 & Fully connected & & $4096$  \\[5pt]
		Fc3 & Fully connected & & $k$  \\[5pt]
		Softmax & & & $k$  \\[5pt]
		\bottomrule
	\end{tabular}
\end{table*}

The size of the convolution kernels is fixed to be $v \times 3 \times 3$ ($view \times height \times width$), which is the same as the 2D CNNs~\cite{simonyan2014very}. It is demonstrated that small spatial receptive of $3 \times 3$ can increase the performance of DNN models in 2D recognition. Therefore, we set the kernel size to be $3 \times 3$ to compute the 2D spatial features in each view image and set the third viewpoint dimension to be $v$ to aggregate spatial correlated features between view images. The number of filters for each convolution layers is 64, 128, 256, 256, 512, 512, 512, and 512, respectively. We add padding to both spatial and views dimension in all convolution layers, so that the size of feature maps remain constant after these layers.

For the pooling layers, to preserve the 2D spatial features in the single-view images, we set the kernel size to be $1 \times 2 \times 2$ with stride of $ 1 \times 2 \times 2 $ in the first pooling layer. In other words, we apply 2D spatial max-pooling on each view image. Apart from the first pooling layer, the remaining pooling layers implement 3D max-pooling with kernel size of $2 \times 2 \times 2$ and stride of $2 \times 2 \times 2$. Therefore, the size of the output feature maps is scaled-down by a factor of 32 ($2^{5}$) compared with the origin input. Meanwhile, the viewpoint dimension is also scaled-down by a factor of 16 ($2^{5-1}$), as shown in TABLE~\ref{para}.

\section{Experiment}
\label{experiment}

\subsection{Experimental setup}
\textbf{Dataset.} We evaluate our MV-C3D model on the 3D ModelNet Benchmark~\cite{wu20153d}. It is a comprehensive collection of 3D CAD models, which contains 127,915 models divided into 662 different categories. As shown in TABLE~\ref{data}, two subsets of ModelNet are widely used, which are ModelNet10 with 4,899 object instances in 10 categories and ModelNet40 with 12,311 object instances in 40 categories. Both of them are fully labeled and used in many state-of-the-art researches~\cite{wang2017dominant, su2015multi, maturana2015voxnet, han2019seqviews2seqlabels, yavartanoo2018spnet}. The datasets also provide both the training and testing sets. For example, ModelNet10 has 3,991 training and 908 testing samples and ModelNet40 has 9,843 training and 2,468 testing samples. We use the default settings in our experiments.
\newline

\textbf{\noindent Training detail.} We perform experiments on a machine with NVIDIA TITAN X Pascal GPU, Intel Core i7-6700K CPU, and 32GB RAM. Our proposed model is coded in the Tensorflow~\cite{abadi2016tensorflow} platform, which is a popular deep learning library from Google.

The neural network is trained using Adam~\cite{kingma2014adam} optimization. The initial learning rate is set to be 0.0001 and divided by 10 every 20 epochs during the training. The loss function $L_{total}$ is cross-entropy with $L_2$ weight regularization as shown in the following equation:
\begin{equation}
L_{total} = -\frac{1}{n}\left [ \sum_{i=1}^{n}\sum_{j=1}^{l} \left \{ y^{(i)}= j \right \} \log\hat{y}^{(i)}\right ] + \frac{\lambda }{2m}\sum w^{2},
\end{equation}
where $n$ is the mini-batch size, $l$ is the number of category (e.g., $l=10$ for ModelNet10, and $l=40$ for ModelNet40), $y^{(i)}$ and $\hat{y}^{(i)}$ represent the true label and the prediction score, respectively, $\left \{ \cdot  \right \}$ is the indicator function, ${\lambda }$ is the weighting parameter which is set to 0.0005 empirically, $w$ is the filter parameters initialized with zero-mean Gaussian distribution with standard deviation of 0.05, and $m$ is the total number of hyper-parameters.

In training phase, we divide the default training into training set and validation set in a ratio of 4 to 1. We calculate the validation loss every epoch and stop the training when validation loss converges in 5 epochs ($\delta _{i} < threshold$, $i\in \{1,2,3,4,5\}$), with $\delta _{i}$ defined by
\begin{equation}
\delta _{i} = \frac{Val\_Loss_{i-1} - Val\_Loss_{i}}{Val\_Loss_{i}}.
\end{equation}

\begin{table}[!t]
	\caption{The details of ModelNet sub-dataset.}
	\label{data}
	\begin{tabular}{|c|c|c|c|}
		\hline
		\multirow{2}{*}{\textbf{Name}}  & \multirow{2}{*}{\textbf{Train split}} & \multirow{2}{*}{\textbf{Test split}} & \multirow{2}{*}{\textbf{Total}} \\
		&  &  &    \\
		\hline
		ModelNet10   & 3991  & 908  & \textbf{4899} \\
		\hline
		ModelNet40    & 9843  & 2468  & \textbf{12311} \\
		\hline
	\end{tabular}
\end{table}

\subsection{Exploring viewpoint dimension of kernel}
A small receptive field of $3 \times 3$ convolution kernel is appropriate for 2D spatial feature learning according to the findings in VGGNet~\cite{simonyan2014very}. Thus we fix the spatial dimension of 3D convolution kernel to $3 \times 3$ when only vary the viewpoint dimension to exploit the optimal 3D convolution kernel size. Moreover, we set $N$ to be 12, which is consistent with the existent methods.

During the experiment, we first assume that all convolution kernels have the same viewpoint dimension. Thus, we evaluate 4 different 3D kernel sizes which the viewpoint dimension fixed to 1, 3, 5, and 7 from the first to the eighth convolution layer. Then we set the dimension varying across different convolution layers. For this setting, we test two types of networks with the viewpoint dimension of kernel size in decreasing order and increasing order, respectively, and choose the one of the best performance to compare with other settings. In particularly, we choose 7-5-5-5-3-3-1-1 to represent the decreasing order and 1-1-3-3-5-5-7-7 for increasing order.

The networks are trained on the training sets of ModelNet10 and tested using the testing sets. FIGURE~\ref{kernel} shows the experimental results. 
The size of 3D convolution kernel  which the viewpoint dimension is fixed to 3 gives the best performance. 
Therefore, we use $3 \times 3 \times 3$ kernels in the following experiments. Moreover, an interesting observation is that when viewpoint dimension is equal to 1, the performance is the worst compared with other settings. This is expected since it is essentially equivalent to a 2D convolution kernel and hence, cannot capture multi-view features. This suggests that 3D CNNs can learn spatial correlated features between multi-view images effectively and improve the classification results.

\Figure[t!](topskip=0pt, botskip=0pt, midskip=0pt)[width=.999\linewidth]{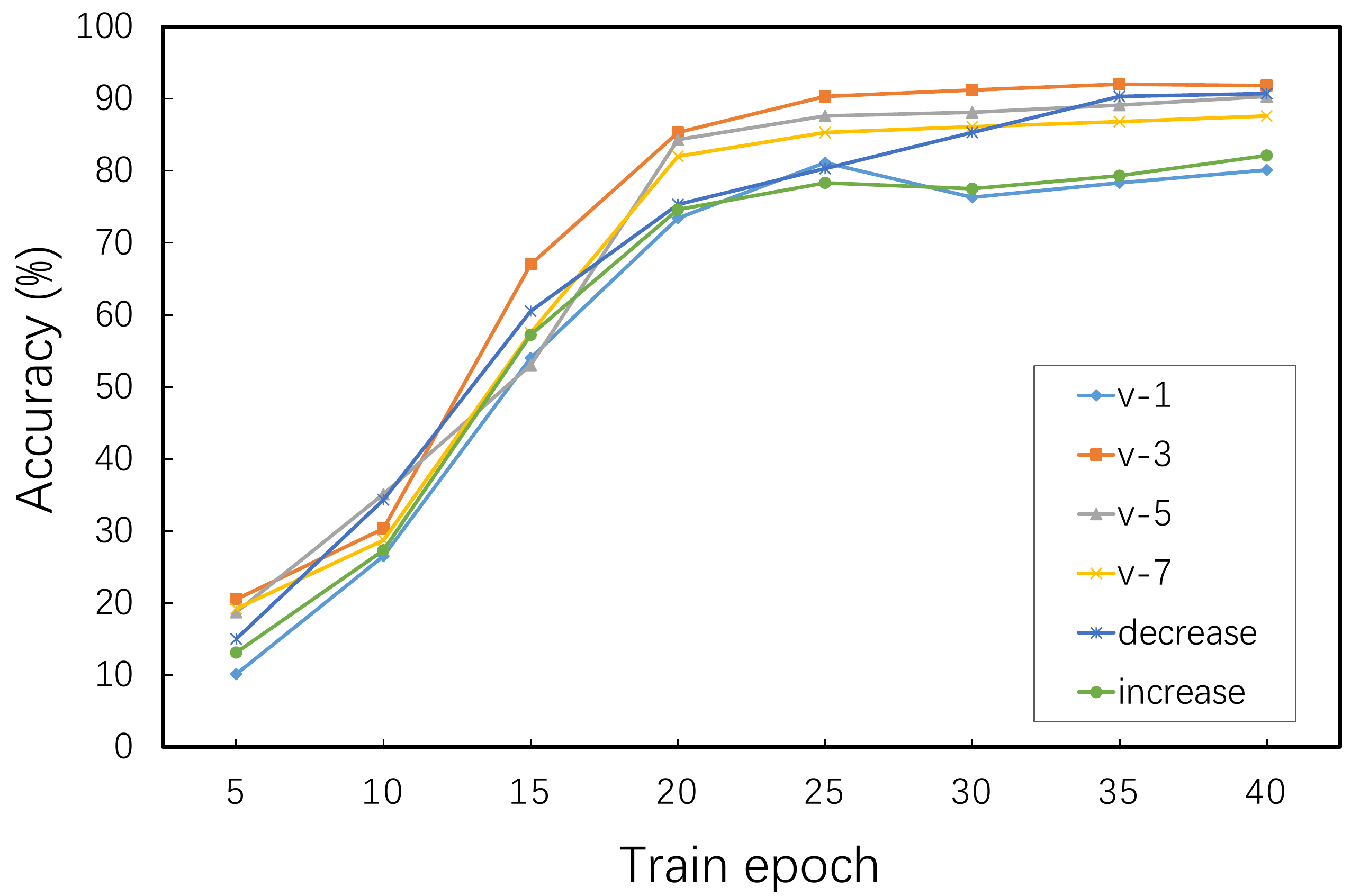}
{Exploring viewpoint dimension of 3D convolution kernel on ModelNet10.\label{kernel}}

\subsection{Over-sampling}
Class imbalance can significantly affect the performance and generalization ability of the models~\cite{buda2017systematic, krawczyk2016learning}. As shown in FIGURE~\ref{d10d40}, the number of instances in each category varies greatly. To eliminate the influence of data bias, we select the object instance which belongs to the fewer categories randomly and designate it as a new instance in the same category. Therefore, the number of instances in each category is balanced. To create a more balanced training set, we increase the number of instances in each category to 500. Thus, our scheme can significantly reduce the imbalanced data problem. After applying our strategies, the classification accuracies on ModelNet10 and ModelNet40 are shown in TABLE~\ref{oversampling}, the model performance is slightly improved.

\begin{table}[!t]
	\caption{Different performances with or without oversampling.}
	\label{oversampling}
	\begin{tabular}{|c|c|c|}
		\hline
		\multirow{2}{*}{\textbf{Method}} & \multirow{2}{*}{\textbf{ModelNet10}} & \multirow{2}{*}{\textbf{ModelNet40}}
		\\
		&  &  \\
		\hline
		No sampling   & 90.5\% & 89.8\% \\
		\hline
		Oversampling  & 91.1\% & 90.1\% \\
		\hline
	\end{tabular}
\end{table}

\Figure[!t](topskip=0pt, botskip=0pt, midskip=0pt)[width=1\linewidth]{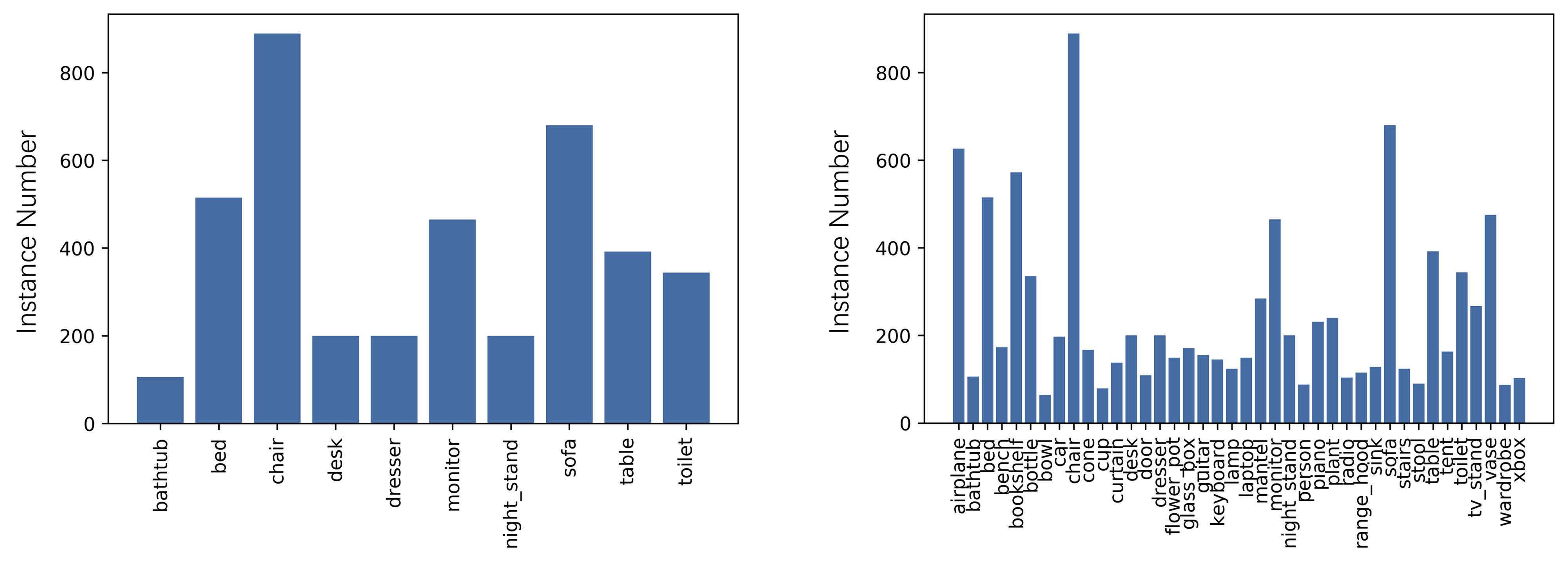}
{The number of instances for each category in ModelNet10 and ModelNet40.\label{d10d40}}

\subsection{Pre-training}
In 2D object classification applications, the model performance can be significantly improved by pre-training on ImageNet~\cite{xuan2018evolving}. Similarly, when pre-trained on ImageNet, the existent 3D multi-view based methods~\cite{su2015multi, hegde2016fusionnet}, which is based on 2D CNNs such as VGG-M~\cite{chatfield2014return}, can also improve the classification accuracy. Unfortunately, our MV-C3D model cannot be pre-trained on ImageNet because of lacking multi-view images. Therefore, for pre-training, we employ the UCF101~\cite{soomro2012ucf101}, which is an action classification dataset collected from Youtube, to pre-train our 3D CNNs. It is demonstrated that 3D CNNs can learn relevance features between different video frames effectively~\cite{tran2015learning}. As shown in TABLE~\ref{pre-training}, our MV-C3D can also derive a better classification accuracy when pre-trained and fine-tuned on UCF101.

\begin{table}[!t]
	\caption{Different performances with different pre-processing methods.}
	\label{pre-training}
	\begin{tabular}{|l|c|c|c|c|}
		\hline
		\multicolumn{1}{|c|}{\multirow{2}{*}{}} & \multirow{2}{*}{\textbf{Pre-training}} & \multirow{2}{*}{\textbf{Oversampling}} & \multirow{2}{*}{\textbf{ModelNet10}} & \multirow{2}{*}{\textbf{ModelNet40}} \\
		\multicolumn{1}{|c|}{}  & & &  & \\ 
		\hline
		1 & $\times$ & $\times$ & 90.5\% & 89.8\% \\ 
		\hline
		2 & $\times$ & $\checkmark$ & 91.1\% & 90.1\% \\ 
		\hline
		3 & $\checkmark$ & $\times$ & 91.7\% & 91.0\% \\ 
		\hline
		4 & $\checkmark$ & $\checkmark$ & 92.0\% & 91.5\% \\ 
		\hline
	\end{tabular}
\end{table}

\subsection{Effect of the number of view images}
\label{analysis}
In this section, we explore the impact of the number of view images on the classification performance. FIGURE~\ref{numberofview-change} shows the performance of our MV-C3D with the number of view images varying from 1 to 36 on ModelNet40. The performance is poor when the number of view images is below 4 because of the lack of sufficient spatial correlated features. With the increasing number, the classification accuracy improves rapidly. Our MV-C3D technique achieves 89.3\% classification with only 10 views and 93.2\% with 16 views. The performance converges to 93.9\% ($\pm$ 0.1\%) with more than 20 views.

\Figure[!t](topskip=0pt, botskip=0pt, midskip=0pt)[width=0.999\linewidth]{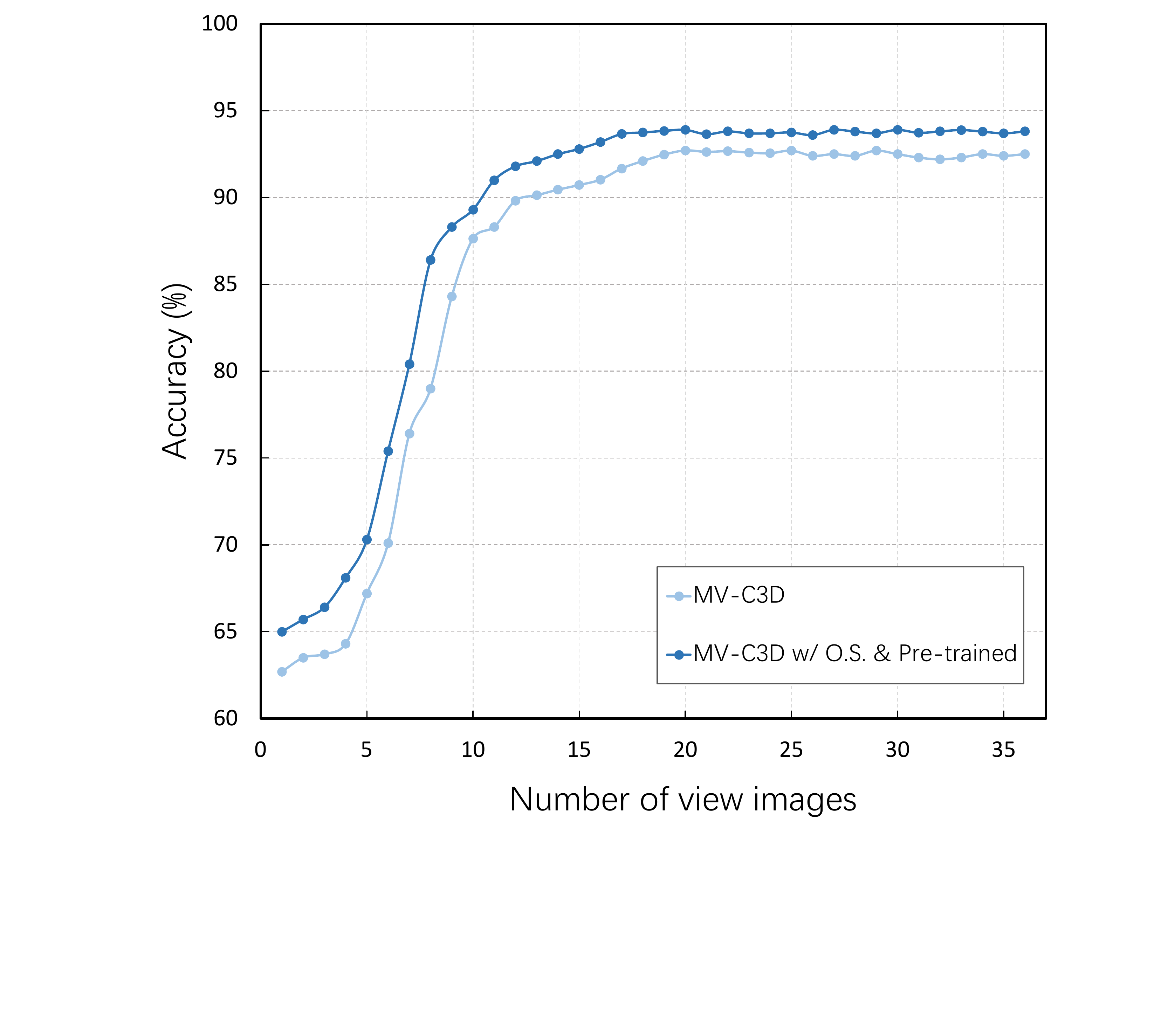}
{Classification result versus number of view images on ModelNet40.\label{numberofview-change}}

\subsection{Experiment on ModelNet}
We compare our MV-C3D technique with other methods based on varying input modalities (e.g., voxel, point cloud). As shown in TABLE~\ref{table}, our proposed MV-C3D model achieves an accuracy of 93.9\% and 90.5\% mAP on ModelNet40, which outperforms all the other methods based on voxel and point cloud. Meanwhile, for multi-view input, it has an improvement of 3.8\% and 11\% compare with MVCNN technique on the classification and retrieval tasks, respectively. For multi-modal methods, our method outperforms all the other methods except for the Spherical Projections~\cite{cao20173d} technique, which is 0.3\% better. However, their approach requires depth information, which is impractical and requires more resources to process. Moreover, MV-C3D achieves the best result of 94.5\% on classification task in ModelNet10.

\begin{table*}[ht]
	\centering
	\caption{Comparison of classification Accuracy and retrieval mean Average Precision (mAP).}
	\label{table}
	\begin{tabular}{lccccc}
		\toprule
		\multirow{3}[4]{*}{\textbf{Method}} & \multicolumn{1}{c}{\multirow{3}[4]{*}{\textbf{Input Modality}}} & \multicolumn{2}{c}{\textbf{ModelNet40}} & \multicolumn{2}{c}{\textbf{ModelNet10}} \\
		\cmidrule{3-6}          &       & \multicolumn{1}{p{6.5em}}{\textbf{Classification}} & \multicolumn{1}{p{4.39em}}{\textbf{Retrieval}} & \multicolumn{1}{p{6.5em}}{\textbf{Classification}} & \multicolumn{1}{p{4.39em}}{\textbf{Retrieval}} \\
		&       & \textbf{(Accuracy)} & \textbf{(mAP)} & \textbf{(Accuracy)} & \textbf{(mAP)} \\
		\midrule
		Beam Search~\cite{xu2016beam} & \multirow{6}[2]{*}{Voxel} & 81.26\% & -     & 88\%  & - \\
		VoxNet~\cite{maturana2015voxnet} &       & 83\%  & -     & 92\%  & - \\
		3D-GAN~\cite{wu2016learning} &       & 83.3\% & -     & 91\%  & - \\
		LightNet~\cite{zhi2017lightnet} &       & 86.9\% & -     & 93.39\% & - \\
		FPNN~\cite{li2016fpnn} &       & 88.4\% & -     & -     & - \\
		\multicolumn{1}{l}{MVCNN-MultiRes~\cite{qi2016volumetric}} &       & 91.4\% & -     & -     & - \\
		\midrule
		3D ShapeNets~\cite{wu20153d} & \multirow{6}[2]{*}{Point Cloud} & 77.32\% & 49.23\% & 83.54\% & 68.26\% \\
		PointNet~\cite{garcia2016pointnet} &       & -     & -     & 77.6\%  & - \\
		PointNet++~\cite{qi2017pointnet++} &       & 91.9\% & -     & -     & - \\
		Set-convolution~\cite{ravanbakhsh2016deep} &       & 90\%  & -     & -     & - \\
		Angular Triplet-Center~\cite{li2018angular} &       & -     & 86.11\% & -     & \textbf{92.07\%} \\
		Geo-CNN~\cite{lan2018modeling} &       & 93.9\% & -     & -     & - \\
		\midrule
		DeepPano~\cite{shi2015deeppano} & \multirow{2}[2]{*}{Multi-view} & 77.63\% & 76.81\% & 85.45\% & 84.18\% \\
		MVCNN~\cite{su2015multi} &       & 90.1\% & 79.5\% & -     & - \\
		\midrule
		FusionNet~\cite{hegde2016fusionnet} & Multi-view + Voxel & 90.8\% & -     & 93.11\% & - \\
		PRVNet~\cite{you2018pvrnet} & Multi-view + Point Cloud  & 93.6\% & \textbf{90.5\%} & -     & - \\
		Multiple Depth~\cite{zanuttigh2017deep} & Multi-view + Depth & 87.8\% & -     & 91.5\% & - \\
		DSCNN~\cite{wang2017dominant} & Multi-view + Depth & 93.8\% & -     & -     & - \\
		Spherical Projections~\cite{cao20173d} & Multi-view + Depth & \textbf{94.24\%} & -     & -     & - \\
		\midrule
		MV-C3D (proposed) & Multi-view & 93.9\% & \textbf{90.5\%} & \textbf{94.5\%} & 91.4\% \\
		\bottomrule
	\end{tabular}
\end{table*}

\subsection{Replicability}

To demonstrate the replicability of our method, we have repeated experiment on ModelNet10 for 20 trials. FIGURE~\ref{Replicable} shows the accuracy curve with error band. The accuracies have high variance at the beginning of training. However, with the increasing of train epoch, the results converge and are stable at 94.5\% ($\pm$ 0.15\%) after 30 training epochs. This result suggests that our method has good replicability.

\Figure[ht](topskip=0pt, botskip=0pt, midskip=0pt)[width=0.999\linewidth]{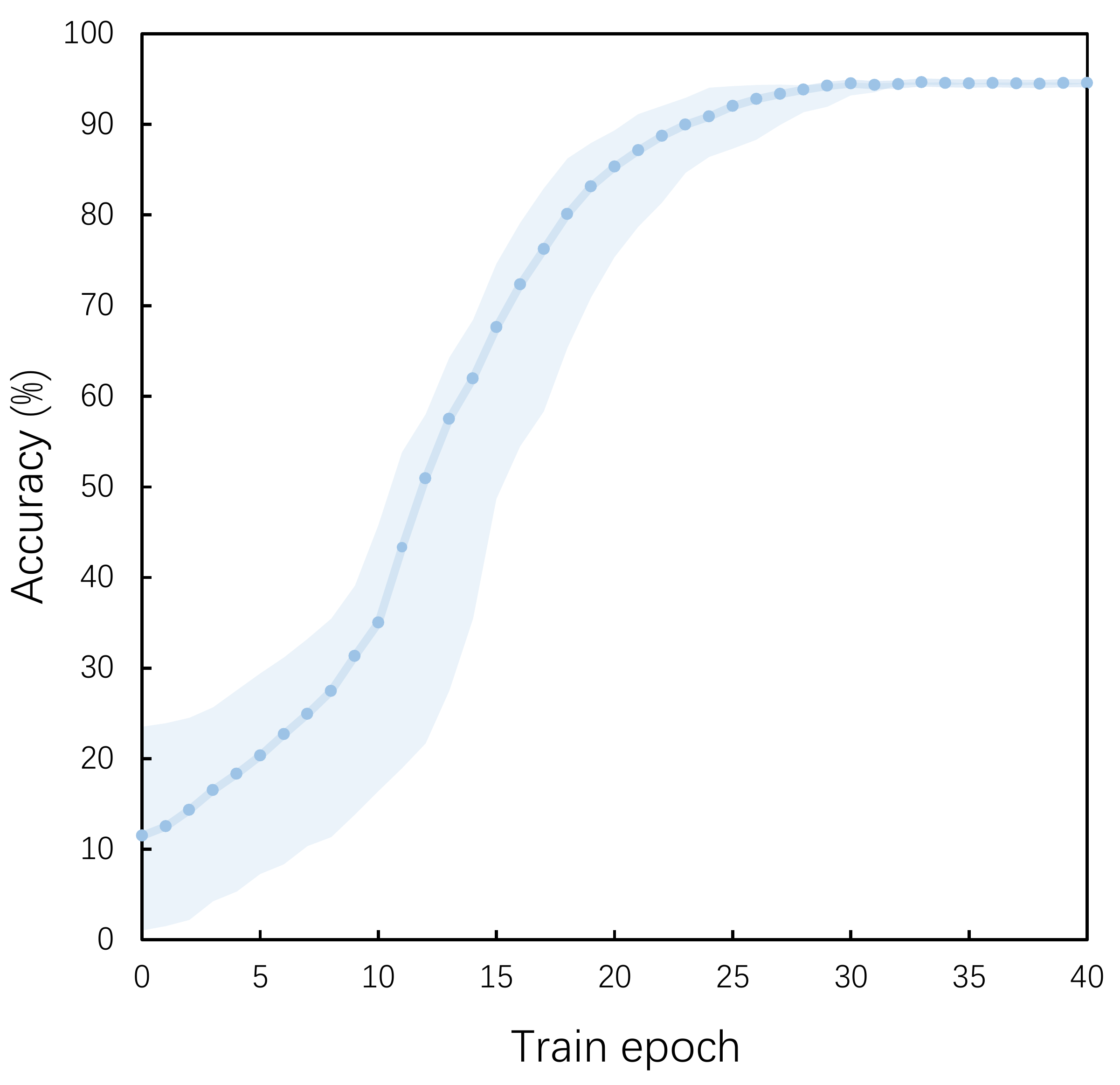}
{Results on ModelNet10 with error band.\label{Replicable}}

\subsection{Visualization of learned features}
For a better understanding of how MV-C3D works, we provide visualization of learned features by using the method from~\cite{zeiler2014visualizing}. FIGURE~\ref{visualization} shows the deconvolution of one learned feature map of the $conv5\_b$ layer. We can see that MV-C3D focuses on empennage in all view images in the first example. The second focuses on empennage and airfoil simultaneously. The third focuses on chair legs and the forth focuses on chair handle. The fact that during the learning process, images from different angles focusing on the same feature suggests that MV-C3D can capture correlated features between multi-view images effectively.

\Figure[!h](topskip=0pt, botskip=0pt, midskip=0pt)[width=1\linewidth]{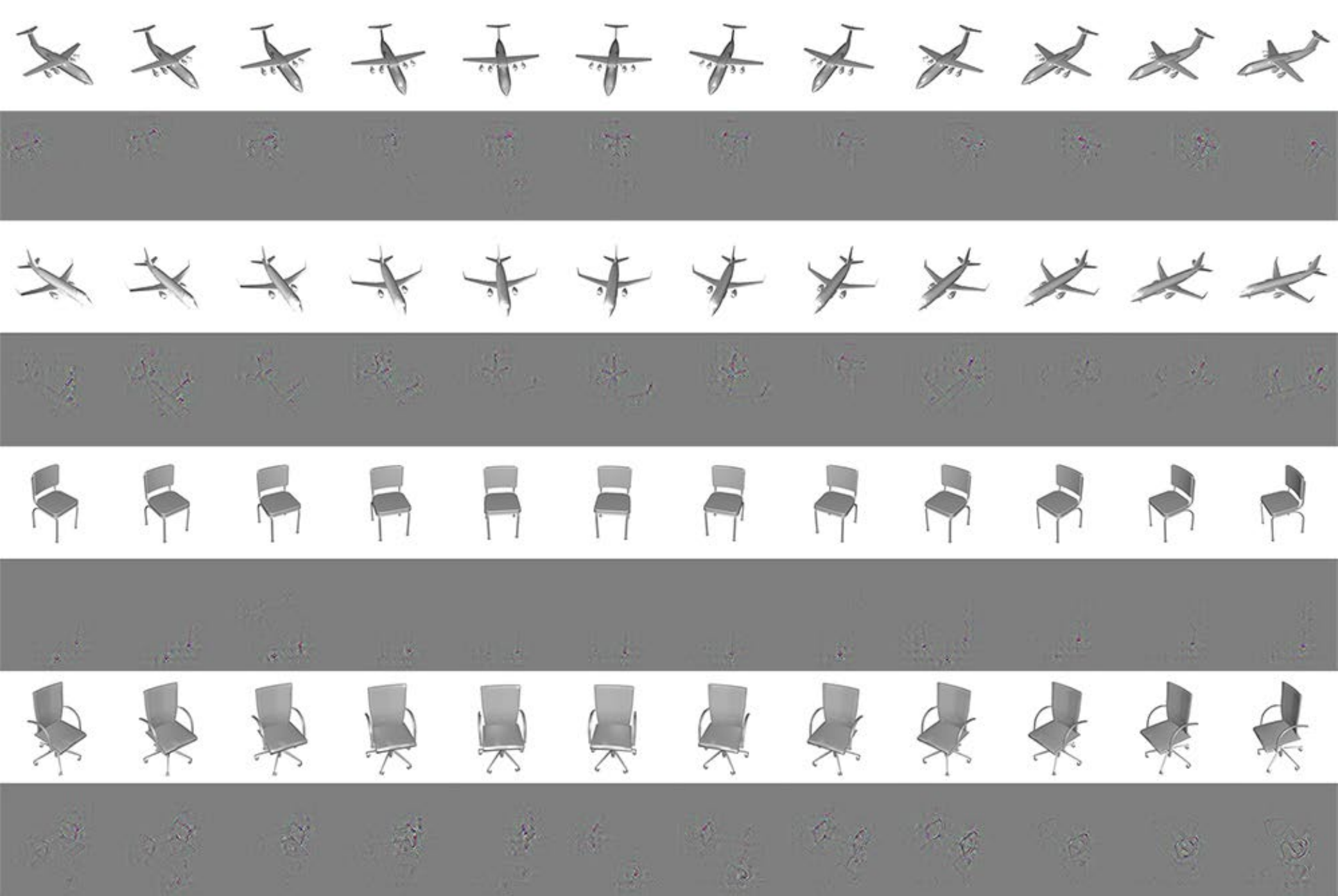}
{Visualization of learned features between multi-view images. In the first example, the feature focuses on empennage in all view images. The second focuses on empennage and airfoil simultaneously. The third focuses on chair leg and the forth focuses on chair handle. \label{visualization}}

\subsection{Comparison with Multi-view based methods}
In this section, we compare our technique with the multi-view based method MVCNN~\cite{su2015multi} and multi-modal based method DSCNN~\cite{wang2017dominant} on ModelNet40 classification task. As shown in TABLE~\ref{comparison}, MV-C3D achieves 91.4\% accuracy by taking all view images with interval 30{\textdegree} ($12\times 30$\textdegree), which outperform MVCNN 0.9\%, but is slightly worse 0.8\% compared to DSCNN, using the same input setting. We believe that the correlated feature information between multi-view is weak when they are highly dissimilar because of the large interval. However, when we take 12 contiguous view images with interval 10{\textdegree} ($12\times 10$\textdegree) as input, the performance is increased to 91.9\% while other methods decreased. With the increasing number of views, the performance of our MV-C3D improves and achieves 93.9\% when the number of views is 20, which outperforms other methods with the same input setting. The reason may be that the prior multi-view based methods mainly rely on global contour information while our MV-C3D only needs the correlated information between multi-view images. This is an advantage when MV-C3D is applied in real-word scenarios where objects are always captured with a limit angle instead of omnibearing.

\begin{table}[ht]
	\centering
	\caption{The comparison between MV-C3D and other multi-view based methods in different input settings on ModelNet40.}
	\label{comparison}
	\begin{tabular}{cccc}
		\toprule
		\textbf{Method} & \multicolumn{1}{p{6.61em}}{\textbf{View Interval}} & \textbf{\#Views} & \textbf{Accuracy} \\
		\midrule
		\multirow{5}[4]{*}{MVCNN} & $30^{\circ}$    & 12    & 89.5\% \\
		\cmidrule{2-4}          & \multirow{4}[2]{*}{$10^{\circ}$} & 8     & 80.1\% \\
		&       & 12    & 82.7\% \\
		&       & 16    & 84.1\% \\
		&       & 20    & 85.3\% \\
		\midrule
		\multirow{5}[4]{*}{DSCNN} & $30^{\circ}$    & 12    & \underline{92.2\%} \\
		\cmidrule{2-4}          & \multirow{4}[2]{*}{$10^{\circ}$} & 8     & 87.6\% \\
		&       & 12    & 90.3\% \\
		&       & 16    & 91.4\% \\
		&       & 20    & 92.1\% \\
		\midrule
		\multirow{5}[4]{*}{MV-C3D} & $30^{\circ}$    & 12    & 91.4\% \\
		\cmidrule{2-4}          & \multirow{4}[2]{*}{$10^{\circ}$} & 8     & 90.5\% \\
		&       & 12    & 91.9\% \\
		&       & 16    & 93.2\% \\
		&       & 20    & \textbf{93.9\%} \\
		\bottomrule
	\end{tabular}
\end{table}

\subsection{Experiment on 3D rotated image dataset}
In this section, we test our technique on a 3D rotated real image dataset "Multi-view Images of Rotated Objects (MIRO)"~\cite{kanezaki2018rotationnet}. In the previous experiments, we assume that the viewpoints are uniformly distributed along a circle. However, in real-world applications, objects are often observed with arbitrary directions, which is more close to MIRO. MIRO consists of 120 object instances in 12 categories, and each instance has 160 images (10 different elevation angles and 16 different azimuth angles) captured from different viewpoints approximately equally distributed in the spherical space. FIGURE~\ref{example} shows an example of object and the corresponding multi-view images. We randomly select 12 contiguous views (both in elevation direction and azimuth direction) as input to test our model, which is trained on the ModelNet40 dataset. The accuracy on each category is reported in TABLE~\ref{miro-acc}. In almost all cases, MV-C3D outperforms other multi-view based methods. This suggests that MV-C3D is accurate, robust, and more practical.

\Figure[!h](topskip=0pt, botskip=0pt, midskip=0pt)[width=0.999\linewidth]{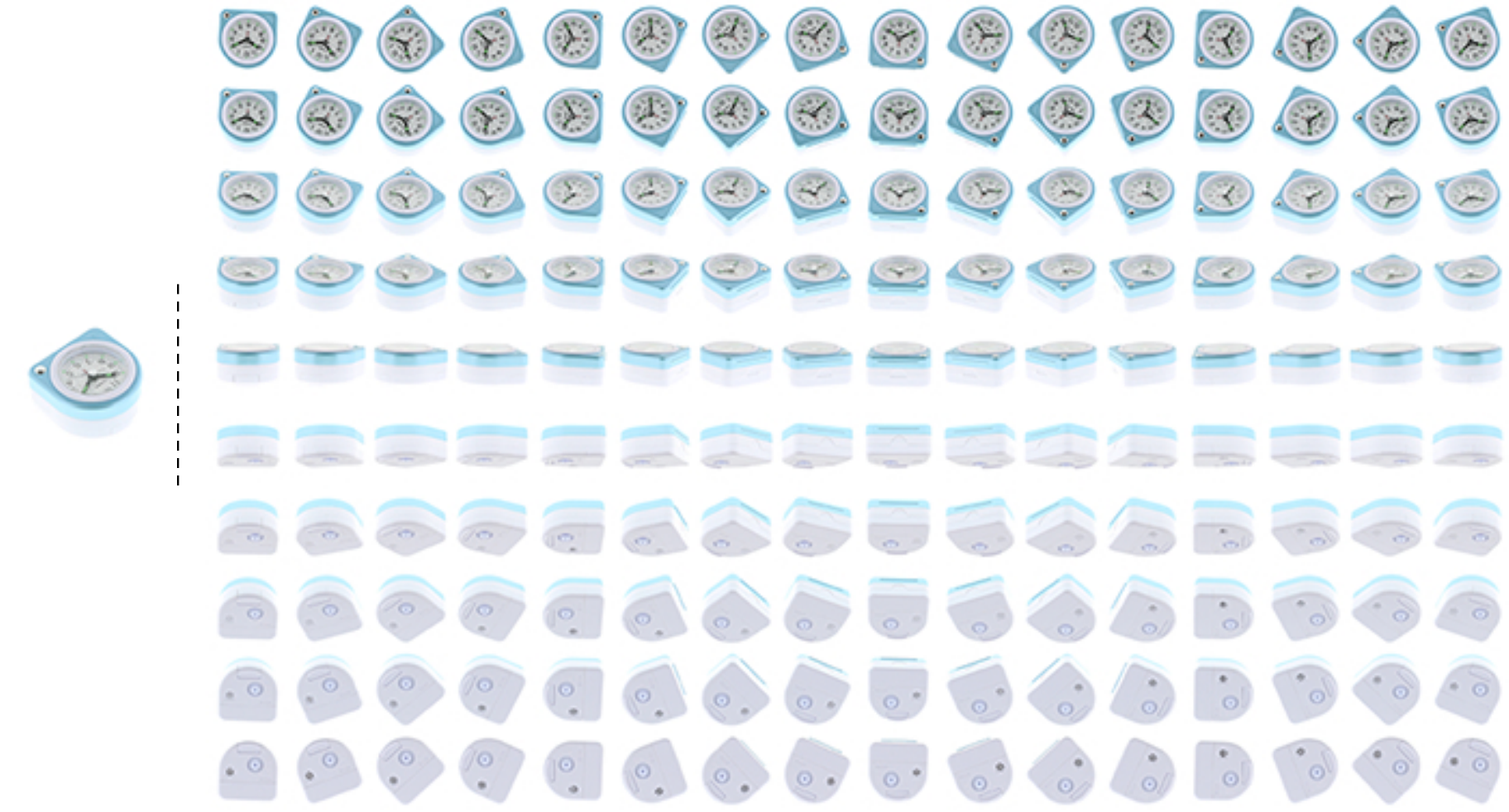}
{A clock exemplar and the multi-view images in MIRO dataset. \label{example}}

\begin{table*}[h]
	\centering
	\caption{Classification accuracy on each category in MIRO.}
	\label{miro-acc}
	\begin{tabular}{cccccccccccccc}
		\toprule
		\multirow{2}{*}{\textbf{Method}} & \multirow{2}{*}{\textbf{bus}} & \multirow{2}{*}{\textbf{car}} & \multirow{2}{*}{\textbf{cleanser}} & \multirow{2}{*}{\textbf{clock}} & \multirow{2}{*}{\textbf{cup}} & \textbf{head-}  & \multicolumn{1}{l}{\multirow{2}{*}{\textbf{mouse}}} & \multicolumn{1}{l}{\multirow{2}{*}{\textbf{scissors}}} & \multicolumn{1}{l}{\multirow{2}{*}{\textbf{shoe}}} & \multicolumn{1}{l}{\multirow{2}{*}{\textbf{stapler}}} & \multicolumn{1}{l}{\multirow{2}{*}{\textbf{sunglasses}}} & \textbf{tape}   & \multicolumn{1}{l}{\multirow{2}{*}{\textbf{Mean}}} \\
		&                               &                               &                                    &                                 &                               & \textbf{phones} & \multicolumn{1}{l}{}                                & \multicolumn{1}{l}{}                                   & \multicolumn{1}{l}{}                               & \multicolumn{1}{l}{}                                  & \multicolumn{1}{l}{}                                     & \textbf{cutter} & \multicolumn{1}{l}{}                               \\
		\midrule
		MVCNN                            & 80\%                          & 70\%                          & 90\%                               & 90\%                            & \textbf{100\%}                & 70\%            & 80\%                                                & 60\%                                                   & 90\%                                               & \textbf{100\%}                                        & 80\%                                                     & 90\%            & 83.3\%                                             \\
		DSCNN                            & \textbf{90\%}                 & 80\%                          & \textbf{100\%}                     & 90\%                            & \textbf{100\%}                & 80\%            & 80\%                                                & 70\%                                                   & 90\%                                               & 90\%                                                  & \textbf{90\%}                                            & 90\%            & 87.5\%                                             \\
		MV-C3D                           & \textbf{90\%}                 & \textbf{90\%}                 & \textbf{100\%}                     & \textbf{100\%}                  & \textbf{100\%}                & \textbf{90\%}   & \textbf{90\%}                                       & \textbf{80\%}                                          & \textbf{100\%}                                     & 90\%                                                  & \textbf{90\%}                                            & \textbf{100\%}  & \textbf{93.3\%}            \\
		\bottomrule                       
	\end{tabular}
\end{table*}

\subsection{Ablation study}
\textbf{Ablation study on convolution pattern.} The prior work employs the 2D convolution to extract feature independently. To evaluate the 3D convolution operation in MV-C3D, we build the same neural network as MV-C3D but use the 2D convolution operation instead. As shown in FIGURE~\ref{2d-independently}, each view image is calculated using individual 2D convolution kernels. TABLE~\ref{2d-3d-conv} shows the performance of different convolution filters on ModelNet40. 3D convolution outperform 2D convolution significantly, which demonstrates the importance of 3D convolution.
\newline

\Figure[!h](topskip=0pt, botskip=0pt, midskip=0pt)[width=0.999\linewidth]{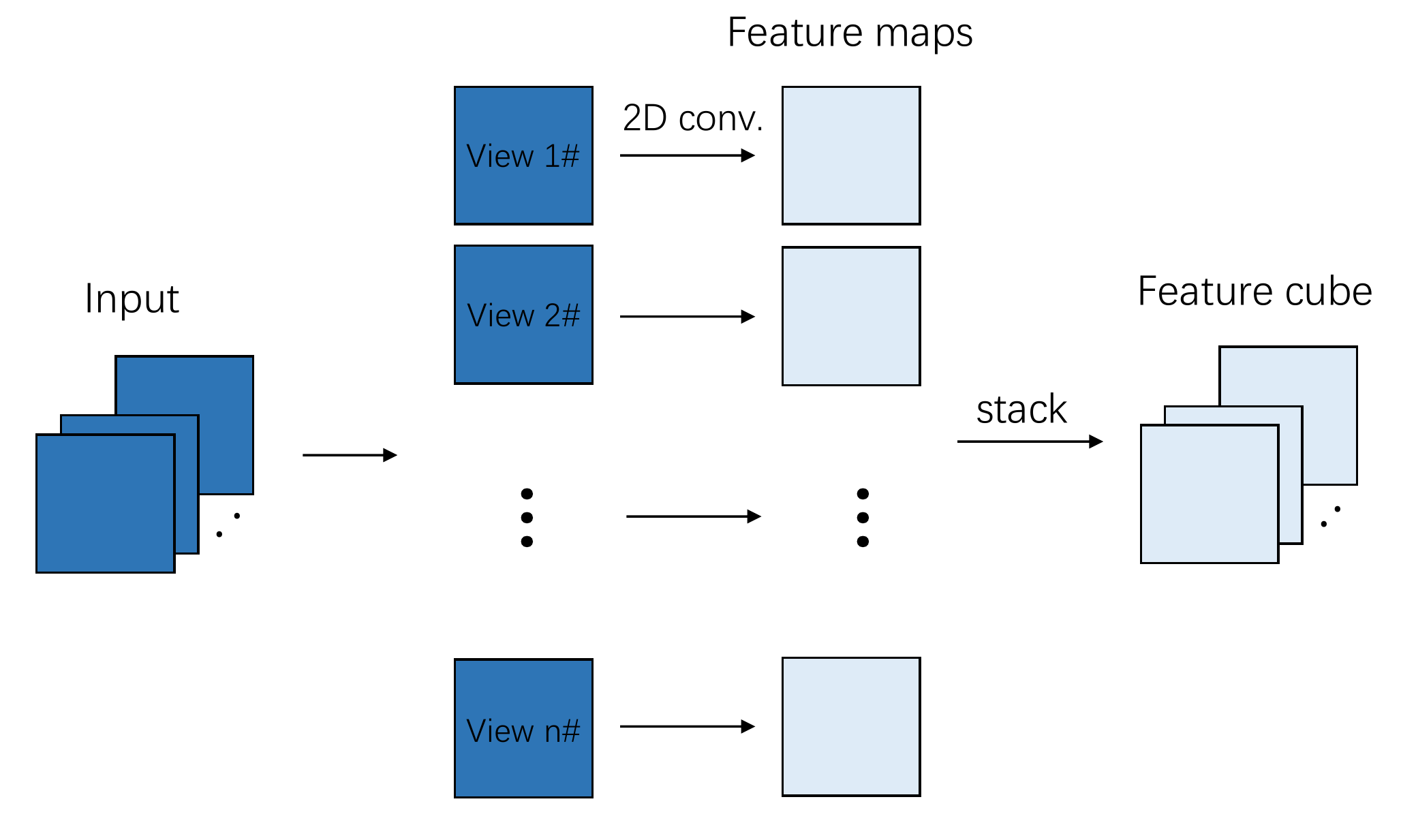}
{Extract features on multi-view images by 2D convolution operation independently. \label{2d-independently}}

\begin{table}[!h]
	\centering
	\caption{Ablation study on convolution pattern. Classification results on ModelNet with different convolution pattern.}
	\label{2d-3d-conv}
	\begin{tabular}{ccc}
		\toprule
		\textbf{Method} & \textbf{ModelNet10} & \textbf{ModelNet40} \\
		\midrule
		2D conv. & 81.8\% & 75.3\% \\
		3D conv. & \textbf{94.5\%} & \textbf{93.9\%} \\
		\bottomrule
	\end{tabular}
\end{table}

\textbf{\noindent Ablation study on model complexity.} Most of the parameters in the MV-C3D model come from the last three fully connected layers. The parameter number can be estimated as $ d_{1}*d_{2} + d_{2} * d_{3} + d_{3} * N_{Classes} $, where $d_{1}$ is determined by the channel numbers and the size of feature maps in the last 3D max-pooling layer,  $N_{Classes} $ is the number of categories, $d_{2}$, $d_{3}$ are the dimensions of the second and third fully connected layer, respectively. We increase $d_{2}$ and $d_{3}$ from 1024  to 4096 to estimate the impact of model complexity. The experimental results are shown in TABLE~\ref{complexity}. For "MV-C3D-S" model, $d_{2}$, $d_{3}$ are set to 1024. For "MV-C3D-M" model, both are 2048. We obverse that the increasing model complexity does not improve performance much. Increasing the number of input view images has a much more significant impact.

\begin{table}[!h]
	\centering
	\caption{Comparison of models with varying complexity.}
	\label{complexity}
	\begin{tabular}{cccc}
		\toprule
		\textbf{Method} & \textbf{Model Size} & \textbf{\#Views} & \textbf{Accuracy} \\
		\midrule
		\multirow{3}[2]{*}{MV-C3D-S} & \multirow{3}[2]{*}{142M} & 8     & 82.0\% \\
		&       & 12    & 88.7\% \\
		&       & 16    & \textbf{92.7\%} \\
		\midrule
		\multirow{3}[2]{*}{MV-C3D-M} & \multirow{3}[2]{*}{186M} & 8     & 82.9\% \\
		&       & 12    & 89.3\% \\
		&       & 16    & \textbf{93.0\%} \\
		\midrule
		\multirow{3}[2]{*}{MV-C3D} & \multirow{3}[2]{*}{299M} & 8     & 83.7\% \\
		&       & 12    & 91.9\% \\
		&       & 16    & \textbf{\underline{93.2\%}} \\
		\bottomrule
	\end{tabular}
\end{table}

\section{Conclusion}
\label{conclusion}
In this paper, we propose MV-C3D, which is a multi-view based 3D convolutional neural network and can perform 3D objects classification using multi-view images which
are captured from only partial angles with less range.  MV-C3D can effectively learn 3D object representations by using 3D convolution layers and max-pooling layers to aggregate the spatial correlated features of different viewpoint images. Experiments on the ModelNet10 and ModelNet40 benchmarks show that MV-C3D outperform the state-of-the-art multi-view based methods by using only RGB images partial viewpoints which can easily be captured by surveillance cameras or moving cameras. Furthermore, the outstanding results on a real image dataset MIRO suggest that our technique can be applied in real-world multi-view classification task. In the future work, we plan to explore different architectures to further reduce the parameters of the 3D convolution based model while maintaining the accuracy of classification.

\bibliographystyle{IEEEtran}
\bibliography{REFS}

\EOD

\end{document}